\documentclass[letterpaper]{article} 
\usepackage{aaai23}  
\usepackage{times}  
\usepackage{helvet}  
\usepackage{courier}  
\usepackage[hyphens]{url}  
\usepackage{graphicx} 
\usepackage{natbib}  
\usepackage{caption} 
\usepackage{algorithm}
\usepackage{algorithmic}

\usepackage{newfloat}
\usepackage{listings}
\DeclareCaptionStyle{ruled}{labelfont=normalfont,labelsep=colon,strut=off} 
\floatstyle{ruled}
\newfloat{listing}{tb}{lst}{}
\floatname{listing}{Listing}
\usepackage{mmstyle}
\usepackage{amsmath}
\usepackage{multirow}
\usepackage{booktabs}
\usepackage{colortbl}
\usepackage{pifont}
\usepackage[caption=false]{subfig}
\usepackage{makecell}
\usepackage{marvosym}
\usepackage[font=small]{caption}
\usepackage{adjustbox}
\usepackage{diagbox}
\usepackage{xcolor}
\definecolor{light-gray}{gray}{0.92}
\definecolor{dark-gray}{gray}{0.50}

\newcommand{\g}[1]{\textcolor{dark-gray}{#1}}
\newcommand{\bd}[1]{\textbf{#1}}

\newcommand{\cmark}{\ding{51}}
\newcommand{\xmark}{\ding{55}}
\usepackage[switch]{lineno} 

\title{SEPT: Towards Scalable and Efficient Visual Pre-Training}
\author{
    Yiqi Lin\textsuperscript{\rm 1}\thanks{This work was done during the internship at SenseTime when his affiliation is Sun Yat-Sen University.}, 
    Huabin Zheng\textsuperscript{\rm 2}, 
    Huaping	Zhong\textsuperscript{\rm 2}, Jinjing Zhu\textsuperscript{\rm 1}, 
    Weijia Li\textsuperscript{\rm 3}\equalcontrib, 
    Conghui	He\textsuperscript{\rm 2},\\
    Lin	Wang\textsuperscript{\rm 1, 4}\equalcontrib
}
\affiliations{
    \textsuperscript{\rm 1}AI Thrust, Information Hub, HKUST (Guangzhou), Guangzhou, China
    \textsuperscript{\rm 2}SenseTime Research\\
    \textsuperscript{\rm 3}Sun Yat-Sen University
    \textsuperscript{\rm 4}Department of Computer Science and Engineering, HKUST, Hong Kong, China
    \{ylin933, jzhu706\}@connect.hkust-gz.edu.cn, \{zhenghuabin,zhonghuaping,heconghui\}@sensetime.com
    \\liweij29@mail.sysu.edu.cn, 
    linwang@ust.hk
}

\usepackage{bibentry}

\begin{document}

\maketitle


\begin{abstract}
Recently, the self-supervised pre-training paradigm has shown great potential in leveraging large-scale unlabeled data to improve downstream task performance.
However, increasing the scale of unlabeled pre-training data in real-world scenarios requires prohibitive computational costs and faces the challenge of uncurated samples.
To address these issues, we build a task-specific self-supervised pre-training framework from a data selection perspective based on a simple hypothesis that pre-training on the unlabeled samples with similar distribution to the target task can bring substantial performance gains.
Buttressed by the hypothesis, we propose the first yet novel framework for \textbf{S}calable and \textbf{E}fficient visual \textbf{P}re-\textbf{T}raining (\textbf{SEPT}) by introducing a retrieval pipeline for data selection.
SEPT first leverage a self-supervised pre-trained model to extract the features of the entire unlabeled dataset for retrieval pipeline initialization.
Then, for a specific target task, SEPT retrievals the most similar samples from the unlabeled dataset based on feature similarity for each target instance for pre-training.
Finally, SEPT pre-trains the target model with the selected unlabeled samples in a self-supervised manner for target data finetuning.
By decoupling the scale of pre-training and available upstream data for a target task, SEPT achieves high scalability of the upstream dataset and high efficiency of pre-training, resulting in high model architecture flexibility.
Results on various downstream tasks demonstrate that SEPT can achieve competitive or even better performance compared with ImageNet pre-training while reducing the size of training samples by \textit{one magnitude without resorting to any extra annotations}.

\end{abstract}

\section{Introduction}
\label{sec:intro}

\begin{figure}[t!]
    \centering
    \includegraphics[width=0.9\linewidth]{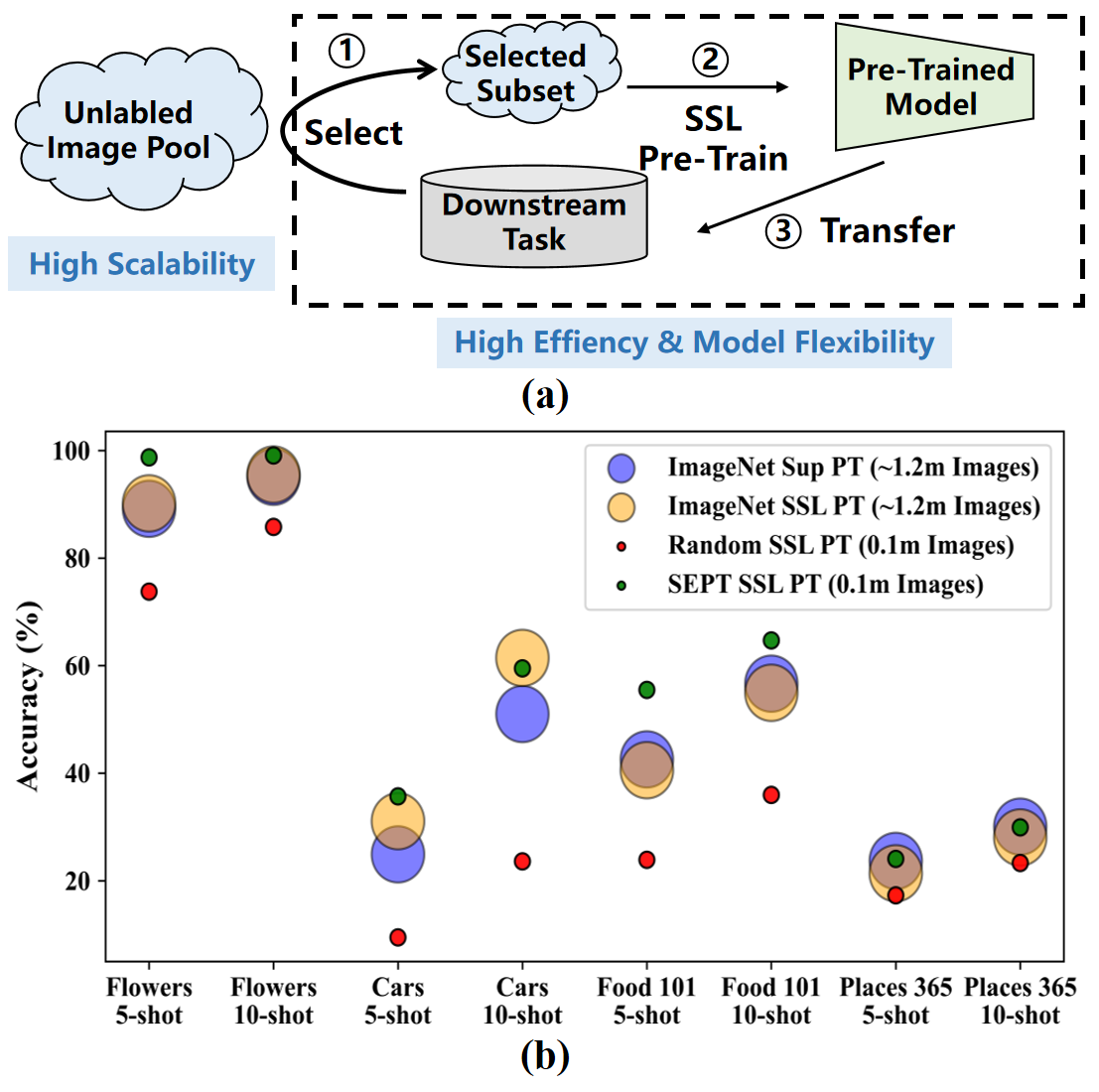}
    \vspace{-12pt}
    \caption{
    (a) The general pipelines of our proposed SEPT.
    (b) The down-stream task classification accuracy of SEPT with 0.1m images and ImageNet~1k baselines. The ratio denotes  the number of pre-training samples under the same training epoch setting.
    }
    \label{fig:teaser}
    \vspace{-10pt}
\end{figure}

To reduce the demand for collecting large-scale labeled data for each target application, supervised pre-training on large-scale datasets, \eg, ImageNet~\cite{deng2009imagenet} and then finetuning on the target tasks have become a successful and standard paradigm in many applications~\cite{he2017mask,long2015fully,sun2017revisiting}.
To avoid such expensive annotation costs,
self-supervised learning (SSL) methods~\cite{doersch2015unsupervised,zhang2016colorful,noroozi2016unsupervised,komodakis2018unsupervised,kingma2013auto,bao2021beit} have shown that models can be trained with freely available supervision from raw data itself.
Recent advances of SSL methods~\cite{caron2021emerging,he2022masked,zhou2021ibot} can achieve comparable or even superior performance to their supervised pre-training version in solving the downstream task.

Several works~\cite{goyal2021self,kolesnikov2020big,pham2021meta,dumoulin2021comparing} suggest that scaling up the pre-training dataset can bring continuous advancement of the state-of-the-art (SoTA) performance on the downstream tasks.
Despite the promising prospects in collecting the unlabeled data, the scalability of unlabeled datasets for SSL methods still suffers from three aspects of challenges in real-world applications.

Firstly, linearly increasing pre-training computation overhead limits the flexibility of model architecture as different application scenarios may request the model with different architectures and scales; thus, it is prohibitively expensive to perform large-scale pre-training for each customized model.
Secondly, training on the entire massive unlabeled data raises the risk of introducing potential bias~\cite{caron2019unsupervised,tian2021divide} because the unlabeled samples collected from the real-world scenario might be low-quality or without specific semantic information.
Lastly, some studies~\cite{ngiam2018domain,ge2017borrowing} point out the redundancy in the large-scale pre-training for downstream tasks by showing that models pre-trained on a subset can achieve comparable even better to the ones pre-trained on the entire dataset. Therefore, some unlabeled samples can be inessential to a downstream task in self-supervised learning.

An intuitive way to address these issues is to decouple the scales of the pre-training dataset by performing the data selection.
Accordingly, in the supervised pre-training pipelines, NDS~\cite{yan2020neural} and SNDS~\cite{cao2021scalable} have been proposed to select the relevant subset according to the downstream task data for supervised pre-training using a data recommendation system. 
\textit{\textbf{However, existing research seldom explored such a problem in self-supervised pre-training.}}

In this work, we explore an alternative to the existing pre-training paradigm and propose a novel self-supervised Scalable and Efficient visual Pre-Training (\textbf{SEPT}) framework (See Fig.~\ref{fig:teaser}(a)). It aims at optimizing the pre-training efficiency while maintaining the scalability of the pre-training dataset scale.
SEPT is based on a hypothesis that \textit{\textbf{training self-supervised learning methods on a subset having similar distribution to the downstream target task should improve performance on the target task}}, inspired by the domain adaptation theory~\cite{ben2010theory,ganin2015unsupervised}.

Supported by the hypothesis, we use the target dataset to perform the instance search via feature similarity for collecting a relatively small task-specific self-supervised pre-training dataset.
Specifically, SEPT consists of three steps: retrieval pipeline initialization, task-specific instance search, and task-specific self-supervised learning.
SEPT first builds a retrieval pipeline using a self-supervised pre-trained model on a subset of the dataset.
Notably, the retrieval pipeline can be easily reused for different target tasks.
Intuitively, the retrieval model serves as a distribution discrepancy metric to minimize the distribution gap between pre-training and the target dataset.
In task-specific instance search, SEPT uses the retrieval model to search the most similar samples for each task sample from the entire upstream dataset to construct a small subset according to the computational constraints.
Finally, SEPT self-supervised pre-train a target model with the retrieved unlabeled subset and then finetune the target model on the task data.

We conduct experiments on \textbf{seven} classification and \textbf{three} detection tasks with limited labeled samples.
Compared with ImageNet~1k pre-training baselines, our SEPT achieves \textit{\textbf{competitive or better performance}} on classification tasks, Flowers, Food101, Stanford Cars, and Places365 with only 100k unlabeled images for pre-training and few shot images for finetuning (See Fig.~\ref{fig:teaser}(b)).

\section{Related Work}
\label{sec:related}
\vspace{-3pt}
\subsection{Self-Supervised Representation Learning}
\vspace{-3pt}
Self-supervised representation learning has shown promising results in model pre-training with freely available supervision from raw data recently~\cite{jing2020self,liu2021self}.
Early works mainly focus on designing handcrafted pretext tasks~\cite{doersch2015unsupervised,zhang2016colorful,noroozi2016unsupervised,komodakis2018unsupervised} using prior knowledge.
More recent works can be categorized in discriminative~\cite{dosovitskiy2014discriminative,bachman2019learning,he2020momentum,chen2020simple,grill2020bootstrap} or generative~\cite{kingma2013auto,xie2021simmim,bao2021beit,he2022masked} fashion.
In the discriminative fashion, contrastive methods~\cite{dosovitskiy2014discriminative,bachman2019learning,he2020momentum,chen2020simple} force the representation of different views of the same image closer and push representations of views from different images away, which achieves comparable performance to its counterpart of supervised pre-training.

We notice that the models pre-trained with contrastive learning have strong instance discriminative power.
Therefore we exploit such ability for building the retrieval pipeline to help downstream tasks find visually similar samples from the general data pool.
Although it is easy to collect massive unlabeled images for self-supervised pre-training, seldom work studies the scalability of computation overhead on large-scale datasets.

\vspace{-6pt}
\subsection{Data Selection for Pre-Training}
\vspace{-3pt}
In the context of image pre-training, there are also several works dedicated to improving the performance~\cite{ngiam2018domain,ge2017borrowing} or efficiency~\cite{yan2020neural,cao2021scalable} from the perspective of selecting the appropriate training data instead of the entire dataset.
\cite{ge2017borrowing} greedily select the most similar categories from the source dataset to be used for pre-training using a proposed similarity metric between the source and target categories.
\cite{ngiam2018domain} proposes to use a model pre-trained on the source dataset to obtain pseudo labels for target images and uses the pseudo to re-weight the source examples.
NDS~\cite{yan2020neural} represents source datasets with the mixture-of-experts model and employs it to perform data search by finding the dataset with similar behavior to the mixture-of-experts.
SNDS~\cite{cao2021scalable} proposes to use intermediary datasets to train mixture-of-experts for indexing the growing scale of the source dataset, which can avoid re-index the source dataset when adding new data.

Despite their promising results, the scalability of source datasets in these supervised pre-training methods is still limited by the extensive human labeling \cite{cao2021scalable} and heavy computation cost on upstream dataset \cite{ngiam2018domain,ge2017borrowing,yan2020neural}.
Differently, our work explores a new self-supervised pre-training paradigm with a highly scalable unlabeled dataset and efficient pre-training computation overhead on a task-specific subset.

\section{Methodology}

\begin{figure*}[t]
    \centering
    \includegraphics[width=\linewidth]{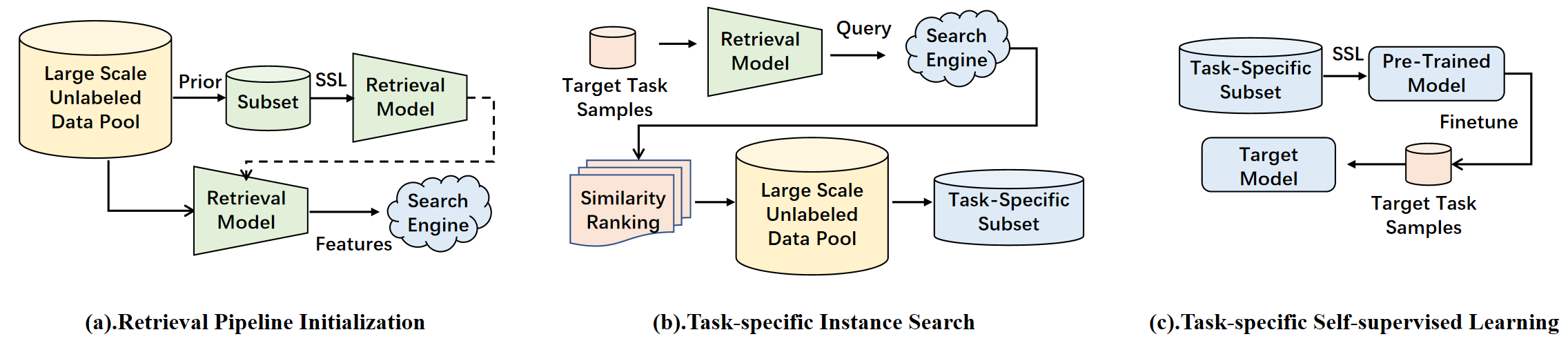}
    \vspace{-18pt}
    \caption{\textbf{Framework Overview}. SEPT consists of three steps: \textbf{a)} retrieval pipeline initialization, \textbf{b)} task-specific instance search and \textbf{c)} task-specific self-supervised learning.
    SEPT first trains a self-supervised model on a subset from a large-scale unlabeled data pool. Then, given a few task data, SEPT retrievals the most similar subset. Finally, SEPT self-supervised pre-trains a target model on the selected subset and fine-tunes the pre-trained model with task samples and their annotations.
    }
    \label{fig:framework}
    \vspace{-8pt}
\end{figure*}

\vspace{-4pt}
\subsection{Preliminaries}
\vspace{-4pt}
\subsubsection{Problem Definition.} 
Consider a general unlabeled dataset $\mathcal{U} = {\{x_i\}}^{N}_{i=0}$ where $x_i$ is an image and a labeled task dataset $\mathcal{T}={(x_i, y_i)}_i$ where $x_i$ is an image and $y_i$ is the ground truth, we assume $|\mathcal{U}|\gg|\mathcal{T}|$.
Our goal is to train a model $\theta$ that fits well on the task $\mathcal{T}$ with the help of an unlabeled dataset $\mathcal{U}$ by finding the most informative subset.
Formally, we seek an optimal unlabeled subset $\mathcal{U}^{*}$ that can help enhance the target task performance with the labeled task dataset $\mathcal{T}$:
\begin{equation}
\mathcal{U}^{*}=\underset{\hat{\mathcal{U}} \subseteq \mathcal{U}}{\arg \min } \mathbb{E}_{x \sim \mathcal{T}} \left[\mathcal{L}\left(\boldsymbol{y} | \theta_{\hat{\mathcal{U}} \cup \mathcal{T}}(\boldsymbol{x}) \right)\right]
\end{equation}
where $\theta_{\hat{\mathcal{U}} \cup \mathcal{T}}$ denotes that $\theta$ is trained with the union of selected unlabeled subset $\hat{\mathcal{U}}$ and labeled task data.
Intuitively, the whole learning system does not involve any annotation from a general unlabeled dataset $\mathcal{U}$ which means $\mathcal{U}$ can be simply scaled up by raw data itself without any human prior.

\vspace{-4pt}

\subsubsection{Theoretical Insight.}
SEPT is based on a hypothesis that \textit{\textbf{training self-supervised learning methods on a subset having similar distribution to the downstream target task should improve performance on the target task}}.
Therefore, our research question is also related to the domain adaptation problem~\cite{pan2009survey}, where the general unlabeled dataset can be viewed as a source domain covering various unknown semantics categories while the target domain respects the specific task.
Unlike the existing domain adaptation methods with well-defined source and target datasets, the general dataset in our problem setting is more uncurated and realistic .
From the perspective of domain adaptation, SEPT focus on sampling the informative source samples to proximate the target distribution, \ie, finding the desired source distribution for better domain adaptation.

Given by \cite{ben2010theory}, let $\mathcal{H}$ be a hypothesis space, the generalization errors of a function $f\in \mathcal{H}$ on the target domain $\mathcal{T}$ and the source domain $\mathcal{S}$ as $\epsilon_t$ and $\epsilon_s$, respectively.
Then, for any function $f\in \mathcal{H}$, we have the following generalization bound,
\begin{equation}
  \begin{array}{c}
 \epsilon_t(f)\le \epsilon_s(f)+d_{\mathcal{H}\Delta\mathcal{H}}(\mathcal{S},\mathcal{T})+\epsilon^*
  \end{array} 
\end{equation}
where $d_{\mathcal{H}\Delta\mathcal{H}}(\mathcal{S},\mathcal{T})$ is $\mathcal{H}\Delta\mathcal{H}$-Divergence which measures the discrepancy between the two domains:
\begin{equation}
  \begin{array}{c}
 d_{\mathcal{H}\Delta\mathcal{H}}(\mathcal{S},\mathcal{T})=\sup\limits_{f,f'\in \mathcal{H}}\big|\mathbb{E}_{x\sim S}[f(x)\neq f'(x)]\\~~~~~~~~~~~~~~~~~~~~~~~~~~~~~~~-\mathbb{E}_{x\sim T}[f(x)\neq f'(x)]\big|
  \end{array} 
\end{equation}
The key problem of domain adaptation is to minimize the distribution discrepancy $d_{\mathcal{H}\Delta\mathcal{H}}(\mathcal{S},\mathcal{T})$.
However, in our setting, the source domain $\mathcal{S}$ is a subset of the unlabeled dataset $\mathcal{S} \in P(\mathcal{U})$ requiring to select.
Therefore, we first need to introduce a pre-defined and generalized metric for measuring the distribution discrepancy.
In~\cite{ganin2016domain}, a domain classifier is introduced to measure the distance between two distributions by recognizing the domain category of samples, and the $d_{\mathcal{H}\Delta\mathcal{H}}(\mathcal{S},\mathcal{T})$ is reformulated as follow,
\begin{equation}
  \begin{array}{c}
 d_{\mathcal{H}_p\Delta\mathcal{H}_p}(\mathcal{S},\mathcal{T})
 \leq 2\sup\limits_{h\in \mathcal{H}_d}\big|
\alpha(h)-1|
  \end{array} 
\end{equation}
where $\mathcal{H}_d$ is the hypothesis space of the domain classifier and $\alpha(h)$ is the optimal classifier.
Inspired by the design of domain classifier, we argue that it is possible to find a general domain discriminator for an arbitrary target task.
In practice, SEPT uses a retrieval pipeline based on feature similarity to proximate the domain discrimination process.
Intuitively, for each target instance, SEPT selects the most similar source sample that mostly confuses the general domain discriminator and thus minimizes the $d_{\mathcal{H}\Delta\mathcal{H}}(\mathcal{S},\mathcal{T})$ with a measurement selected by prior knowledge.
After selecting the source samples, SEPT adapts the most straightforward strategy, pre-training and fine-tuning, to perform domain adaptation for simplicity.

\vspace{-6pt}
\subsection{Proposed Framework}
\vspace{-3pt}
\subsubsection{Overview.}
SEPT is a pre-training framework aiming to find a task-relevant subset from a massive unlabeled image pool given a specific task with limited samples.
The framework consists of three steps: retrieval pipeline initialization, task-specific instance search, and task-specific self-supervised learning.
In the first step, we extract and store features of the entire unlabeled dataset $\mathcal{U}$ with a self-supervised model.
In the second step, we perform the instance search based on feature similarity for each sample in $\mathcal{T}$ to aggregate the most similar subset $\hat{\mathcal{U}}$ from $\mathcal{U}$.
In the last step, a downstream model is self-supervised, pre-trained on the selected subset $\hat{\mathcal{U}}$ and then fine-tuned with target data $\mathcal{T}$ only with the task objective.
Fig. \ref{fig:framework} shows the overall pipeline of our proposed method.

\begin{algorithm}[t]
\caption{Task-specific Instance Search}
\label{alg:algorithm}
\textbf{Input}: an unlabeled dataset $\mathcal{U}$, a target dataset $\mathcal{T}$, a budget (number of images) of pre-training dataset $\mathcal{K}$ and a feature extractor $\theta_R$ well trained on subset  $\hat{\mathcal{U}} \subseteq \mathcal{U}$. \\
\textbf{Output}: a task-specific pre-training subset $D_{search}$.\\
\vspace{-12pt}
\begin{algorithmic}[1]
\STATE $\mathcal{F}_u \leftarrow \theta_R(x^*), \forall x^* \in \mathcal{U}$
\FOR{$x_i~~in~~\mathcal{T}$}
\STATE $F_i \leftarrow \theta_R(x_i)$
\STATE $RankedList(x_{i}) \leftarrow Sorted(sim(F_i,F_u))$
\ENDFOR
\STATE $D_{search} \leftarrow \emptyset $
\FOR{$j~~in~~\mathcal{K}$}
\FOR{$x_i~~in~~\mathcal{T}$}
    \IF{$ D_{search} \cap {RankedList(x_{i})[j]}~~is~~\emptyset $}
    \STATE $ D_{search} \leftarrow D_{search} \cup {RankedList(x_{i})[j]} $
    \IF{$ |D_{search}| \geq \mathcal{K} $}
    \STATE Exit
    \ENDIF
    \ENDIF
\ENDFOR
\ENDFOR
\STATE \textbf{return} selected subset $D_{search}$
\end{algorithmic}
\end{algorithm}

\vspace{-4pt}
\subsubsection{Retrieval Pipeline Initialization.}
To be consistent with our self-supervised pre-training framework, we use a self-supervised pre-trained model as the retrieval model for feature extraction.
Theoretically, pre-training a giant model across the entire $\mathcal{U}$ should be the optimal solution.
However, we argue that it is critical to introduce certain prior knowledge for building the retrieval models for two reasons.
Firstly, the retrieval model only needs to be trained once during initialization, which makes the computational cost of initialization not contribute to the marginal cost of indexing or queries and scaling up the data pool.
Secondly, training on the entire massive unlabeled data raises the risk of introducing potential bias to retrieval models because it is impossible to know the visual concept distribution of the data pool.
Recent self-supervised methods~\cite{chen2020mocov2,caron2021emerging,zhou2021ibot} have shown promising instance discrimination on the unlabeled images and provided lots of successful practice on the ImageNet with various architectures.
Therefore, we use ImageNet self-supervised pre-trained model $f$ as the feature extractor for retrieval to avoid the computation cost and the risk of introducing noise.
The retrieval model can be viewed as a replaceable module in our framework and can be integrated with the latest techniques in this area as they emerge. 
In our experiments, we train the feature extractor using ibot~\cite{zhou2021ibot} with ViT-S~\cite{dosovitskiy2020image}.

\vspace{-4pt}
\subsubsection{Task-specific Instance Search.}
As depicted in \textbf{Algorithm 1}, for each image in the task data $x_i \in \mathcal{T}$, we retrieve a set of image $\hat{\mathcal{U}}_i$ from the given general data pool $\mathcal{U}$.
The set $\hat{\mathcal{U}}_i$ represents the top-$K$ similar images to $x_i$ in $\mathcal{U}$. 
Retrieved data for all images $x_i$ are combined to obtain a subset $\hat{\mathcal{U}} = Union(\hat{\mathcal{U}_0}, ..., \hat{\mathcal{U}_i})$.
Considering the computation cost produced during the retrieval on a large-scale data pool,\eg, 150 million in our setting, we keep our retrieval approach as simple as possible by calculating the similarity between two samples.
Specifically, we use a self-supervised retrieval model $f$ to measure the similarity $sim(f(x_i), f(x_u)), x_u \in \mathcal{U}$ between the task sample and unlabeled samples.
SEPT can handle the different budget constraints of the model pre-training by setting the scale of retrieved samples for real-world applications.  
Our data retrieval method only depends on the raw images $x_i$ itself and does not leverage any task-specific information, such as its labels.

\vspace{-4pt}
\subsubsection{Task-specific Self-Supervised Learning.}
After obtaining the retrieval dataset, we conduct self-supervised pre-training on them by optimizing the following objective:
\begin{equation}
    \theta^{*} = \underset{\theta}{\arg \min } \mathbb{E}_{x \sim \hat{\mathcal{U}}} \left[\mathcal{L}_{self}\left(\boldsymbol{x}\right)\right]
\end{equation}
Note that task-specific self-supervised learning is decoupled from the self-supervised retrieval models.
Therefore, the former can use the latest techniques or various backbones in this area as they emerge.

When transferring to the downstream target task, only task data is used for fine-tuning the pre-trained backbone.
\textit{The exploration of improving performance by joint training with retrieval and task data is left for future work}.
Given the task-specific self-supervised models, we train a task model using the pre-training model as initialization with the following loss, formulated as:
\begin{equation}
    \theta^{*} = \underset{\theta}{\arg \min } \mathbb{E}_{x \sim \mathcal{\mathcal{T}}} \left[ \mathcal{L}_{sup}\left(y| \boldsymbol{x}\right)\right]
\end{equation}
\vspace{-4pt}

\vspace{-12pt}
\section{Experiments Setup}
\label{sec:experiments}
In this section, we conduct experiments to quantitatively evaluate SEPT on various target datasets and analyze the reasonableness of our proposed framework.
Our main experiments are evaluated with limited labeled target samples (5-shot/10-shot) for each category as the effectiveness of pre-training diminishes with the growing size of the dataset~\cite{he2019rethinking}.
Moreover, we also verify the generalization ability of SEPT on classification tasks with more labeled data and detection tasks. 

\vspace{-6pt}
\subsection{Datasets and Setup}
\label{sec:setting}
\vspace{-3pt}
\subsubsection{Datasets for Retrieval Pipeline.}
We combine three large-scale datasets, ImageNet-22k(IN22k)~\cite{deng2009imagenet}, INTERN~\cite{shao2021intern} and YFCC-100m~\cite{thomee2016yfcc100m}, to construct an unlabeled data pool with totally 155 million images, called SEPT-155m.
IN22k collected 14.2 million images from 21,841 classes and has a widely used subset ILSVRC2012 (IN1k), which consists of 1.2 million images from 1,000 classes.
The INTERN classification dataset consists of 40m images with more than 115K concepts.
YFCC-100m contains approximately 99.2 million photos.
The default retrieval model is self-supervised pre-trained on IN1k, which can be viewed as a subset of the entire unlabeled data pool.
Note that we do not access any annotation information during the experiments.

\begin{table}[t]
\centering
\begin{adjustbox}{width=0.9\linewidth}
\begin{tabular}{c|c|l|c|cc|c} 
\toprule[1pt]
\multirow{2}{*}{Task Dataset} & \multirow{2}{*}{\begin{tabular}[c]{@{}c@{}}Number \\ Classes\end{tabular}} & \multirow{2}{*}{Scale} & \multirow{2}{*}{\begin{tabular}[c]{@{}c@{}}Cross Super\\ Category\end{tabular}} & \multicolumn{2}{c|}{Training Samples} & \multirow{2}{*}{\begin{tabular}[c]{@{}c@{}}Testing \\ Samples\end{tabular}} \\ 
\cline{5-6}
 &  &  &  & 5-shot & 10-shot &  \\ 
\hline
Flowers102 & 102 & \multirow{4}{*}{Small} & \xmark & 510 & 1020 & 6149 \\
Stanford Cars & 196 &  & \xmark & 980 & 1960 & 8041 \\
Food101 & 101 &  & \xmark & 505 & 1010 & 25250 \\
Place365 & 365 &  & \xmark & 1825 & 3650 & 36500 \\ 
\hline
iNat-P1k & 1000 & \multirow{3}{*}{Large} & \xmark & 5000 & 10000 & 17157 \\
iNat-I1k & 1000 &  & \xmark & 5000 & 10000 & 17598 \\
iNat-M1k & 1000 &  & \cmark & 5000 & 10000 & 18646 \\
\bottomrule[1pt]
\end{tabular}
\end{adjustbox}
\vspace{-8pt}
\caption{Statistical information of target task datasets.}
\label{tab:stat}
\vspace{-8pt}
\end{table}

\vspace{-5pt}
\subsubsection{Datasets of Target Task.}
We provide an extensive evaluation of our approach on seven classification datasets.
We randomly sample 5-shot or 10-shot from each category for all datasets to construct our target tasks.
The first four tasks are built from Flowers\cite{nilsback2008automated}, Stanford Cars\cite{krause20133d}, Food101\cite{bossard2014food}, Places365\cite{zhou2017places}, respectively.
To comprehensively simulate complex tasks on a scale of IN1k (1k categories), the last three tasks, INat-P1k,  INat-I1k, and INat-M1k, are all constructed from iNaturalist 2017\cite{van2018inaturalist} by sampling different 1,000 categories to build a subset.
All categories in INat-P1k/INat-I1k are sampled from the same super-category of Plantae/Insecta.
Differently, categories in INat-M1k are evenly sampled from 13 different super categories.
The statistical information of classification datasets is summarized in Table~\ref{tab:stat}.
\textit{More details about these datasets are summarized in the appendix}.
In each of them, the SEPT’s goal is to improve the performance of its target task by transferring knowledge from a set of relevant unlabeled images in a self-supervised manner.
 
\vspace{-5pt}
\subsubsection{Baselines and Evaluation.}
In this regime, we compare our framework with the random initialization, IN1k supervised pre-training, and IN1k self-supervised pre-training models since they are widely adopted standard practices in various vision applications.
To demonstrate the effectiveness of data selection, we also provide the baseline pre-trained with randomly sampled images from an unlabeled data pool.
For a fair comparison, we evaluate all the pre-training baselines and SEPT under 300 epochs pre-training protocol and report the performance under the same finetuning setting expect different pre-trained model initialization.

\vspace{-7pt}
\subsection{Implementation Details}
\label{sec:imp-detail}
\vspace{-3pt}
\subsubsection{Retrieval Pipeline.}
The retrieval model use ViT-S~\cite{dosovitskiy2020image} pre-trained on IN1k using self-supervised methods ibot\cite{zhou2021ibot} for 800 epochs.
We use the combination of [CLS] token and patch tokens as features for an image with a dimension of 768.
To perform efficient searching, we use milvus~\cite{wang2021milvus} to build the retrieval pipeline.
Specifically, we use IVF\_SQ8H as index type, set the \textit{nlist} of index to 16384 and use 256 \textit{nprobe} for searching.
The images used for feature extraction are resized to 256$\times$256 and then center cropped with 224$\times$224.

\vspace{-5pt}
\subsubsection{Target Task Pre-training.} 
All experiments are conducted on Swin-T ~\cite{liu2021swin}. 
The self-supervised pre-training follows MoBy~\cite{xie2021self} in 300 epochs setting with batch size 512 on 8 Tesla V100 GPUs.
The pre-training adopts AdamW~\cite{loshchilov2018decoupled} with a fixed learning rate of 0.001 and a fixed weight decay of 0.05.
The key queue size is set to 4096, the temperature is set to 0.2, and the drop path rate is set to 0.2.

\vspace{-5pt}
\subsubsection{Target Task Finetuning.}
All finetuning experiments use the same 100-epoch finetuning setting on single Tesla V100 GPU.
In finetuning, we set the batch size to 64 and employ an AdamW optimizer with a base learning rate of 5e-3, weight decay of 0.05, a stochastic depth \cite{huang2016deep} ratio of 0.1 and a layer-wise learning rate decay of 0.9. 
We also adopt a cosine learning rate scheduler with 10-epoch warm-up. 
We follow the same data augmentation used in \cite{xie2021simmim}, including RandAug\cite{cubuk2020randaugment}, Mixup\cite{zhang2018mixup}, Cutmix\cite{yun2019cutmix}, label smoothing\cite{szegedy2016rethinking}, and random erasing\cite{zhong2020random}.

\begin{table}
\centering
\resizebox{\linewidth}{!}{%
\begin{tabular}{c|c|cc|ccc} 
\toprule
\multicolumn{2}{c|}{\multirow{2}{*}{\begin{tabular}[c]{@{}c@{}}Pre-Train \\Samples\end{tabular}}} & Rand Init & IN1k Sup & IN1k SSL & Random & SEPT \\ 
\cline{3-7}
\multicolumn{2}{c|}{} & 0 & 1200k & 1200k & 100k & 100k \\ 
\hline
Datasets & \begin{tabular}[c]{@{}c@{}}Fine-tune \\Samples\end{tabular} & \multicolumn{5}{c}{Top 1 accuracy} \\ 
\hline\hline
\multirow{2}{*}{Flowers} & 5-shot & \g{19.90} & \g{89.17} & 90.18 & 73.77 & \textbf{98.73} \\
 & 10-shot & \g{31.20} & \g{95.12} & 95.54 & 85.80 & \textbf{99.09} \\ 
\hline
\multirow{2}{*}{Stanford Cars} & 5-shot & \g{2.81} & \g{24.92} & 31.09 & 9.48 & \textbf{35.75} \\
 & 10-shot & \g{3.74} & \g{51.01} & \textbf{61.45} & 23.62 & 59.52 \\ 
\hline
\multirow{2}{*}{Food101} & 5-shot & \g{4.78} & \g{42.58} & 40.57 & 23.91 & \textbf{55.59} \\
 & 10-shot & \g{7.48} & \g{56.66} & 54.96 & 35.97 & \textbf{64.70} \\ 
\hline
\multirow{2}{*}{Place365} & 5-shot & \g{2.87} & \g{23.70} & 21.40 & 17.30 & \textbf{24.08} \\
 & 10-shot & \g{5.12} & \g{30.08} & 28.06 & 23.32 & \textbf{29.94} \\ 
\hline\hline
\multirow{2}{*}{iNat-P1k} & 5-shot & \g{2.73} & \g{30.30} & 29.47 & 16.50 & \textbf{32.33} \\
 & 10-shot & \g{5.74} & \g{47.08} & \textbf{46.60} & 29.54 & 45.82 \\ 
\hline
\multirow{2}{*}{iNat-I1k} & 5-shot & \g{1.67} & \g{33.60} & \textbf{30.70} & 13.71 & 28.58 \\
 & 10-shot & \g{4.10} & \g{49.92} & \textbf{46.57} & 26.12 & 38.55 \\ 
\hline
\multirow{2}{*}{iNat-M1k} & 5-shot & \g{2.85} & \g{36.40} & \textbf{31.90} & 15.72 & 24.93 \\
 & 10-shot & \g{5.51} & \g{48.57} & \textbf{43.96} & 25.03 & 33.65 \\
\bottomrule
\end{tabular}
}
\vspace{-8pt}
\caption{Classification results with 100k images pre-training under 5-shot and 10-shot setting on SEPT-155m unlabeled data pool. IN1k Sup, IN1k SSL, and Random denote the IN1k supervised pre-training, IN1k self-supervised pre-training, and self-supervised pre-training with the randomly selected sample.}
\label{tab:100k}
\vspace{-6pt}
\end{table}

\begin{table}[t]
\centering
\resizebox{\linewidth}{!}{%
\begin{tabular}{c|c|cc|cc|cc} 
\toprule[1pt]
\multirow{2}{*}{\begin{tabular}[c]{@{}c@{}}Number of \\ Samples\end{tabular}} & \multirow{2}{*}{Method} & \multicolumn{2}{c|}{Food101} & \multicolumn{2}{c|}{iNat-I1k} & \multicolumn{2}{c}{iNat-M1k} \\ 
\cline{3-8}
 &  & 5-shot & 10-shot & 5-shot & 10-shot & 5-shot & 10-shot \\ 
\hline
\multirow{2}{*}{1200k} & IN1k Sup & \g{42.58} & \g{56.66} & \g{33.60} & \g{49.92} & \g{36.4} & \g{48.57} \\
 & IN1k SSL & 40.57 & 54.96 & 30.70 & 46.57 & 31.90 & 43.96 \\ 
\hline
\multirow{2}{*}{100k} & Random & 23.91 & 35.97 & 13.71 & 26.12 & 15.72 & 25.03 \\
 & SEPT & 55.59 & 64.70 & 28.58 & 38.55 & 24.93 & 33.65 \\ 
\hline
\multirow{2}{*}{500k} & Random & 32.70 & 47.25 & 20.12 & 36.00 & 22.88 & 34.21 \\
 & SEPT & 69.72 & 76.21 & 44.43 & 55.97 & 33.74 & 45.55 \\ 
\hline
\multirow{2}{*}{1000k} & Random & 34.53 & 49.99 & 22.76 & 40.03 & 24.84 & 37.18 \\
 & SEPT & \textbf{72.17} & \textbf{81.01} & \textbf{49.95} & \textbf{59.52} & \textbf{37.19} & \textbf{48.38} \\
\bottomrule[1pt]
\end{tabular}
}
\vspace{-8pt}
\caption{Classification results with different scales of pre-training data. IN1k Sup, IN1k SSL, and Random denote the IN1k supervised pre-training, IN1k self-supervised pre-training, and self-supervised pre-training with randomly selected samples.}
\label{tab:trainscale}
\vspace{-8pt}
\end{table}

\vspace{-4pt}
\subsection{Main Results}
\label{sec:main-results}
\vspace{-3pt}
\subsubsection{Pre-training with 100k Images.}
Table~\ref{tab:100k} shows the main results that compare SEPT in scales of 100k pre-training images and the according baselines.
In conclusion, the pre-training samples selected by SEPT provide helpful knowledge to all target datasets compared to the random sampling baseline.
In all small-scale downstream datasets, SEPT can achieve results that are better than or comparable to the supervised or self-supervised IN1k baselines with 1/12 of the scale of pre-training samples.
As the large-scale downstream dataset with one thousand visual concepts, the results indicate that general knowledge learned from IN1k pre-training still is a plus for these complex tasks.
Moreover, the performance on three large-scale datasets indicates that SEPT has the edge over fine-grained classification tasks.
Specifically, the performance gaps between SEPT and IN1k  baselines on datasets collected from single super categories iNat-P1k and iNat-I1k are relatively small than iNat-M1k, including multiple super categories images.
Note that 100k images might not be sufficient enough to solve such complex tasks with intensive categories.
Therefore, in the next section, we verify our method with more pre-training samples on both small and large-scale downstream datasets.
Moreover, the performance gap between supervised and self-supervised IN1k baseline also varies on different datasets, which indicates that the ground truth also plays a crucial role in transferring knowledge for different downstream tasks.

\vspace{-5pt}
\subsubsection{Scaling Up the Number of Pre-training Images.}
Table~\ref{tab:trainscale} shows the comparison across different scales of pre-training samples.
In the cases shown, the model pre-trained on SEPT outperforms the baseline using random sampling images in all scales of pre-training samples at a large margin.
Remarkably, when SEPT is pre-trained with 1000k selected images, a comparable size to IN1k, outperforms IN1k supervised pre-training baseline around 30\% in Food101 under 5-shot and 10-shot settings.
We observe that SEPT models pre-trained with 500k samples outperform the IN1k self-supervised baseline on three downstream datasets.
In addition, the increasing performance along with the growing scale of pre-training samples provide strong evidence of the scalability of SEPT in the size of pre-training samples.
Moreover, SEPT achieves more performance gains on the fine-grained dataset iNat-I1k than iNat-M1k, indicating that SEPT is a fine-grained friendly solution for downstream tasks.
Interestingly, IN1k supervised pre-training still is a strong baseline for complex tasks like iNat-M1k, whose classes are across multiple super categories.
Nevertheless, our methods can be scaled up by retrieving more unlabeled images.

\begin{table}
\centering
\resizebox{0.9\linewidth}{!}{%
\begin{tabular}{c|c|c|c|cc} 
\toprule[1pt]
\multirow{2}{*}{\begin{tabular}[c]{@{}c@{}}Retrieval\\ Model\end{tabular}} & \multirow{2}{*}{\begin{tabular}[c]{@{}c@{}}Pre-train \\ Setting\end{tabular}} & \multirow{2}{*}{\begin{tabular}[c]{@{}c@{}}Feature\\ Dim\end{tabular}} & \multirow{2}{*}{\begin{tabular}[c]{@{}c@{}}IN1k \\ Linear Val\end{tabular}} & \multicolumn{2}{c}{Food101} \\ 
\cline{5-6}
 &  &  &  & 5-shot & 10-shot \\ 
\hline
ViT-S & \multirow{3}{*}{\begin{tabular}[c]{@{}c@{}}IN1k \\ 800 epochs\end{tabular}} & 768 & 77.9 & 49.68 & 60.99 \\
ViT-B &  & 1536 & 79.5 & 52.25 & 62.60 \\
ViT-L &  & 2048 & 81.0 & 51.10 & 61.43 \\ 
\hline
ViT-B & \multirow{2}{*}{\begin{tabular}[c]{@{}c@{}}IN22k \\ 80 epochs\end{tabular}} & 1536 & 79.0 & 49.83 & 62.15 \\
ViT-L &  & 2048 & \bd{82.3} & \bd{53.64} & \bd{64.66} \\
\bottomrule[1pt]
\end{tabular}
}
\vspace{-8pt}
\caption{Results with different retrieval models under 100K pre-training setting using IN22k as unlabeled data pool.}
\label{tab:ab-retri}
\end{table}

\begin{table}[t]
\vspace{-6pt}
\centering
\resizebox{\linewidth}{!}{%
\begin{tabular}{c|c|c|cc|cc|cc} 
\toprule
\multicolumn{1}{c|}{\multirow{2}{*}{\begin{tabular}[c]{@{}c@{}}Data\\Pool\end{tabular}}} & \multirow{2}{*}{\begin{tabular}[c]{@{}c@{}}Scale\end{tabular}} & \multirow{2}{*}{Method} & \multicolumn{2}{c|}{Food101} & \multicolumn{2}{c|}{iNat-I1k} & \multicolumn{2}{c}{iNat-M1k} \\ 
\cline{4-9}
\multicolumn{1}{c|}{} &  &  & 5-shot & 10-shot & 5-shot & 10-shot & 5-shot & 10-shot \\ 
\hline
\multirow{2}{*}{IN1k} & \multirow{2}{*}{1.2m} & Random & 26.57 & 38.50 & 16.40 & 28.80 & 16.98 & 26.26 \\
 &  & SEPT & 33.71 & 48.80 & 17.96 & 31.44 & 22.39 & 32.19 \\ 
\hline
\multirow{2}{*}{IN22k} & \multirow{2}{*}{14m} & Random & 25.81 & 38.66 & 14.57 & 26.69 & 16.99 & 26.43 \\
 &  & SEPT & 49.68 & 60.99 & 23.44 & 36.24 & 24.44 & 33.34 \\ 
\hline
\multirow{2}{*}{\begin{tabular}[c]{@{}c@{}}IN22k\\+INTERN\end{tabular}} & \multirow{2}{*}{55m} & Random & 26.03 & 38.70 & 15.25 & 28.15 & 17.46 & 26.47 \\
 &  & SEPT & 54.82 & \textbf{64.80} & 27.46 & 38.14 & 24.68 & 33.40 \\ 
\hline
\multirow{2}{*}{\begin{tabular}[c]{@{}c@{}}IN22k+INTERN\\+YFCC-100m\end{tabular}} & \multirow{2}{*}{155m} & Random & 23.91 & 35.97 & 13.71 & 26.12 & 15.72 & 25.03 \\
 &  & SEPT & \textbf{55.59} & 64.70 & \textbf{28.58} & \textbf{38.55} & \textbf{24.93} & \textbf{33.65} \\ 
\hline\hline
\multirow{2}{*}{YFCC-100m} & \multirow{2}{*}{100m} & Random & 21.69 & 33.62 & 10.38 & 22.76 & 14.26 & 23.33 \\
 &  & SEPT & 51.49 & 61.90 & 23.48 & 35.45 & 24.37 & 32.77 \\
\hline
\end{tabular}
}
\vspace{-8pt}
\caption{Results on various scales of unlabeled data pool under 100K pre-training setting. Each data pool is built from a different combination of large-scale datasets.}
\label{tab:poolsize}
\vspace{-8pt}
\end{table}

\vspace{-3pt}
\subsection{Ablation Study}
\label{sec:ab}
\vspace{-3pt}
\subsubsection{Retrieval Models.}
Table~\ref{tab:ab-retri} compares different retrieval methods (i.e., ViT-S, ViT-B and Vit-L) using IN22k as unlabeled data pool.
We find that given the same data pool, the high capacity of the retrieval model trained with more diverse data can usually provide more performance gain.
Specifically, ViT-L trained with IN22k can bring 4\% improvement in both 5-shot and 10-shot settings compared to the ViT-S trained with IN1k.
Our framework can be upgraded using more powerful self-supervised methods trained with higher capacity and a larger dataset scale depending on the budget for building the retrieval pipeline.

\vspace{-5pt}
\subsubsection{Sizes of Retrieval Unlabeled Data Pool.}
To study the performance gains brought from the increasing scale of the unlabeled data pool, we verify our method on four different scales of data pool by combining different datasets.
Results of 100k pre-training settings with different scales of unlabeled data pool are shown in Table~\ref{tab:poolsize}.
Our methods surpass the random sampling at a large margin on all scales of data unlabeled data pool.
Increasing the data pool size can consistently improve performance on most downstream tasks, demonstrating the advantage of scaling up the pre-training dataset.
It also can be observed that the performance gain becomes trivial when we further increase the image pool size from 55 million to 155 million.
To further verify SEPT in a more realistic setting, we provide results only on YFCC-100m, significantly less curated datasets.
The SEPT and random sampling baseline from YFCC-100m are relatively weak compared to others, which indicates that the distribution shift of unlabeled data can raise the risk of decreasing the representation quality in pre-trained models.
Nevertheless, SEPT still can achieve substantial performance compared to a random sampling baseline in YFCC-100m setting.

\begin{figure}[t!]
    \centering
    \includegraphics[width=0.8\linewidth, height=5cm]{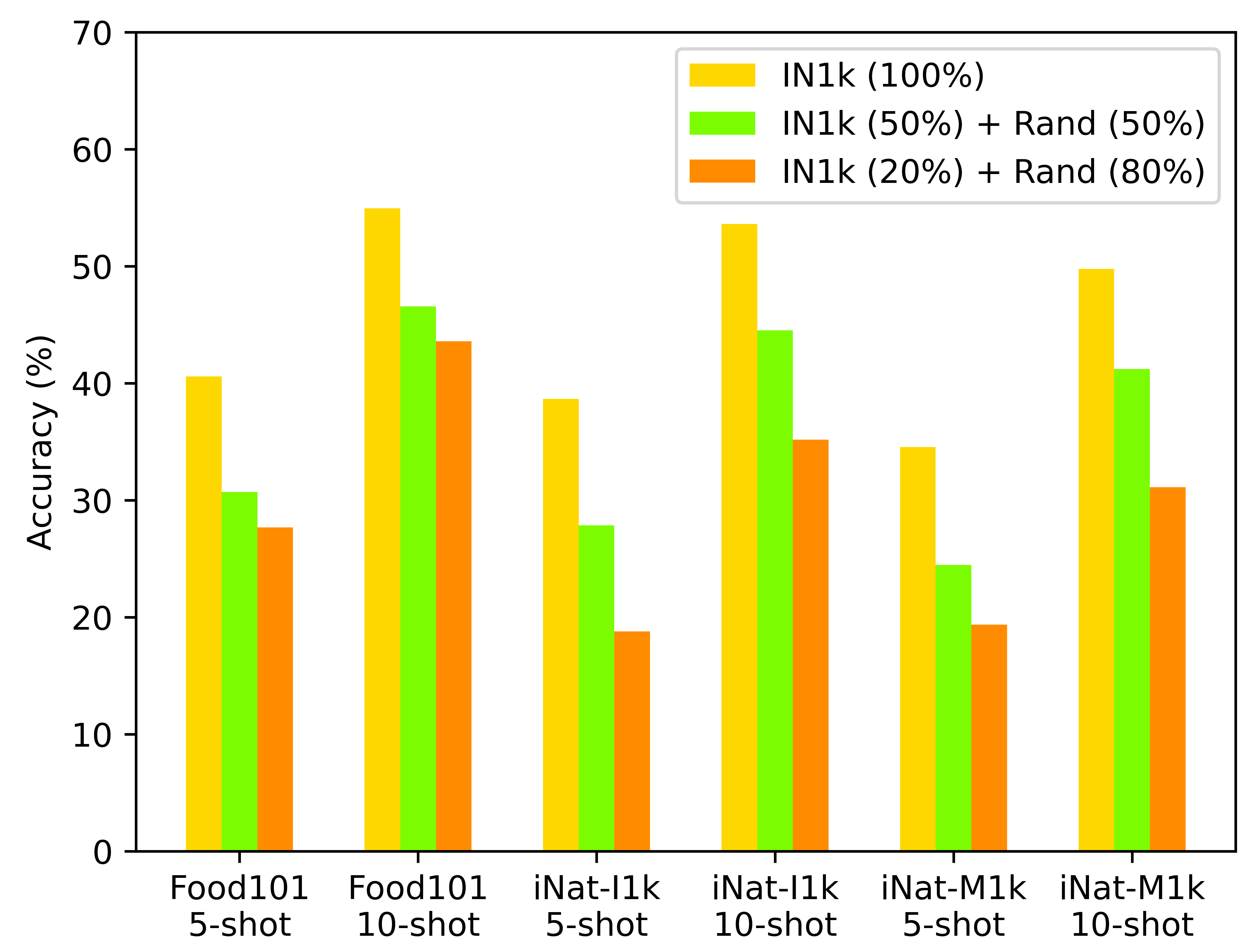}
    \vspace{-8pt}
    \caption{IN1k equivalent self-supervised pre-training from different distribution samples. IN1k 50\%/20\% denote 50\%/20\% pre-training samples are randomly selected from IN1k while remains are randomly sampled from SEPT-155m.}
    \label{fig:distribution_shift}
\end{figure}

\vspace{-3pt}
\subsection{Analysis of Hypothesis and Generalization}
\vspace{-3pt}
\subsubsection{Potential Risk of Distribution Shift.}
Fig.~\ref{fig:distribution_shift} shows that using samples from different distributions can significantly affect downstream performance under IN1k equivalent self-supervised pre-training.
It indicates that applying the current SSL method to different distribution datasets needs a careful design for robustness.
Moreover, it verifies that pre-train on IN1k dataset is a generally transferable and practical solution for pre-training, which can serve as a solid prior for retrieval model initialization.

\vspace{-5pt}
\subsubsection{Redundancy in IN1k.}
\begin{figure}[t!]
    \vspace{-8pt}
    \centering
    \includegraphics[width=0.8\linewidth]{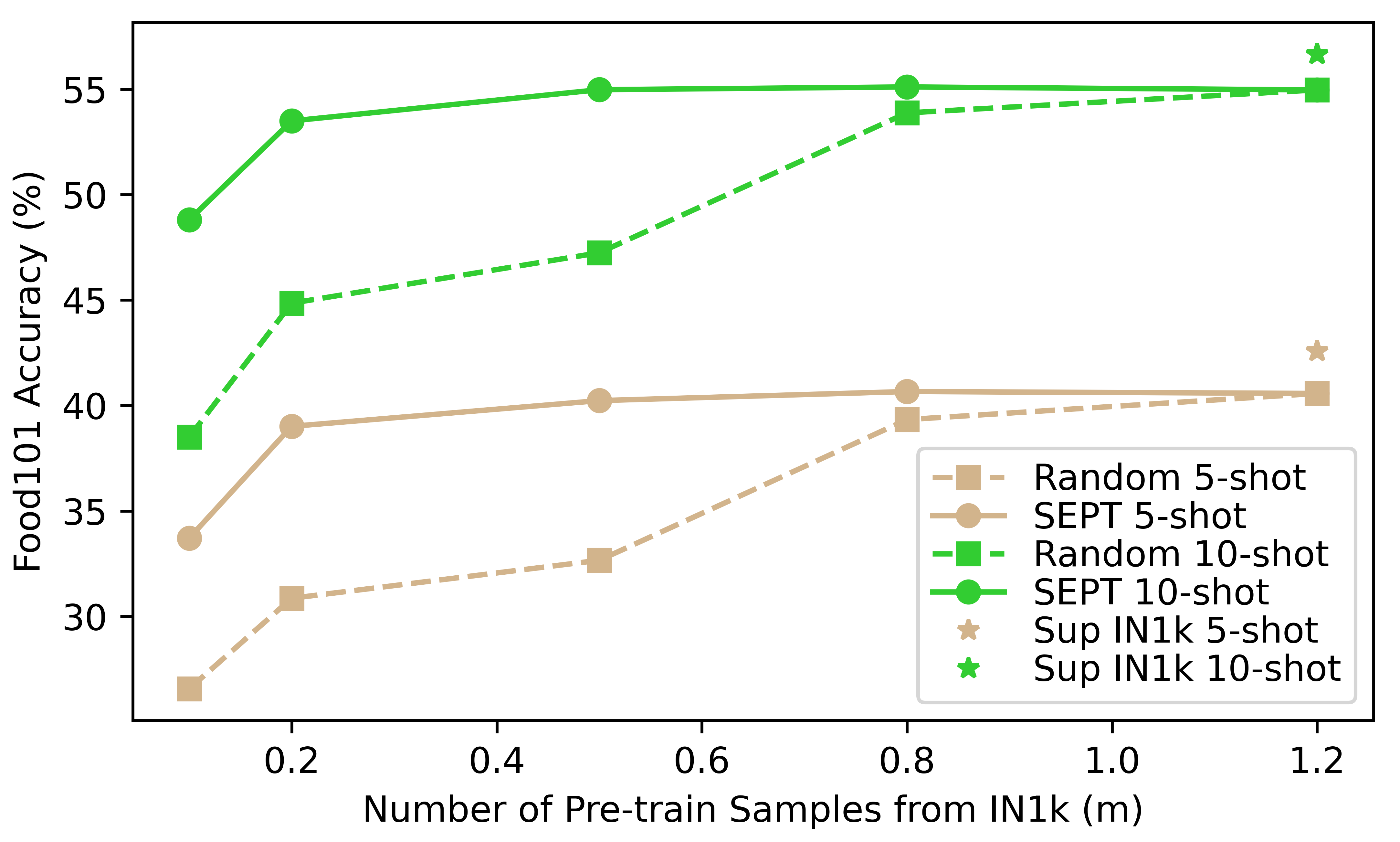}
    \vspace{-8pt}
    \caption{The comparison of using different numbers of IN1k pre-training images in Food101 5-shot and 10-shot tasks.}
    \label{fig:redundancy}
    \vspace{-8pt}
\end{figure}

Fig.~\ref{fig:redundancy} shows results on the Food101 dataset with different proportions of IN1k samples for pre-training. 
SEPT using 500k images for pre-training achieves comparable performance against the full IN1k dataset setting, meaning that not all the pre-training samples are informative for a specific task.
The performance gaps between SEPT and a random sample further verify the superiority of our method across different settings.
\vspace{-5pt}
\subsubsection{Generalization with More Downstream Samples.}
Table~\ref{tab:fullsetting} shows the results using 10-shot SEPT pre-trained models under 50\% and 100\% Food101 dataset (about 70,000 images) finetuning setting.
The results show that SEPT surpasses all the random baselines in different scales of pre-training when using more labeled images, which proves the effectiveness of SEPT more comprehensively.
Consistent with the few shot setting, SEPT pre-trained with 500k images can achieve even better results against IN1k pre-trained baseline.
We also notice that the performance gaps between different pre-trained models will significantly reduce when the model can access more labeled images.
It motivates us to propose the more challenging yet practical few shot setting for effective evaluation.

\begin{table}[t]
\centering
\resizebox{\linewidth}{!}{
\begin{tabular}{c|c|c|c|c|c|c|c|c} 
\toprule
\multirow{2}{*}{\begin{tabular}[c]{@{}c@{}}Pre-train\\Samples\end{tabular}} & \begin{tabular}[c]{@{}c@{}}IN1k\\SSL\end{tabular} & \begin{tabular}[c]{@{}c@{}}Rand\\Init\end{tabular} & Random & SEPT & Random & SEPT & Random & SEPT \\ 
\cline{2-9}
 & 1200k & 0 & \multicolumn{2}{c|}{100k} & \multicolumn{2}{c|}{500k} & \multicolumn{2}{c}{1000k} \\ 
\hline
50\% & 90.08 & 61.64 & 84.34 & 88.11 & 88.57 & 91.74 & 89.39 & \textbf{92.95} \\
100\% & 92.31 & 77.40 & 88.44 & 90.64 & 91.57 & 93.15 & 91.98 & \textbf{93.97} \\
\bottomrule
\end{tabular}
}
\vspace{-8pt}
\caption{SEPT finetuning with more downstream samples on Food101.}
\label{tab:fullsetting}
\vspace{-8pt}
\end{table}

\vspace{-5pt}
\subsubsection{Generalization on Detection.}
To investigate the generation ability of SEPT on other tasks, we conduct experiments on object detection under similarly limited labeled images setting.
The detection datasets include CityScapes~\cite{cordts2016cityscapes}, VOC~\cite{Everingham10} and LogoDet-3k~\cite{wang2022logodet}.
For all detection tasks, we randomly sample 1,000 images from the original training set for our experiments.
In LogoDet-3k task, we create a miniature version of the dataset, called Logo-100, with a 100 class subset for our experiments.
Given the annotation of bounding boxes, we evaluate SEPT with two retrieval strategies, including using the whole image as a query and instance as a query.
When using an instance for retrieval, each instance will be cropped and resized to a fixed size for feature extraction.
Table~\ref{tab:det} shows the results of detection.
Our method outperforms random sampling baseline under 100k and 300k pre-training settings.
The results indicate that using an instance as a query can bring more performance gain from SEPT.
In CityScapes task, SEPT can achieve comparable performance to IN1k baselines under 300k pre-training.
In VOC and Logo dataset, IN1k pre-training still is a strong baseline.
On the one hand, all the categories in VOC task are collected in IN1k, resulting in the strong generalization ability of IN1k pre-training models.
On the other hand, we also find that some logos are very rare in our unlabeled data pool by visualizing the retrieval result, which suggests that IN1k pre-training still is a decent option when SEPT can not validly collect enough relevant pre-training data.

\begin{table}[t]
\centering
\resizebox{0.8\linewidth}{!}{%
\begin{tabular}{c|c|c|c|c}
\toprule[1pt]
Method & \begin{tabular}[c]{@{}c@{}}Pre-train\\Samples\end{tabular} & CityScapes & VOC & Logo-100 \\ 
\hline
Random Init & 0 & \g{11.5} &  \g{13.9} & \g{33.8} \\
\hline
IN1k Sup & 1200k & \g{33.2} & \g{67.5} & \g{69.5} \\
IN1k SSL & 1200k & 33.4 & \bd{67.8} & \bd{69.9} \\ 
\hline
Random & 100k & 26.3 & 47.0 & 55.3 \\
SEPT-Image & 100k & 29.9 & 55.6 & 64.3 \\ 
SEPT-Instance & 100k & 31.4 & 57.4 & 64.6 \\ 
\hline
Random & 300k & 30.0 & 60.2 & 61.5\\
SEPT-Image & 300k & 33.4 & 60.6 & 65.9 \\
SEPT-Instance & 300k & \bd{34.9} & 63.3 & 65.2 \\

\bottomrule[1pt]
\end{tabular}
}
\vspace{-6pt}
\caption{\small{Results for object detection using 1,000 samples for each dataset. Performance are measured in mAP.}}
\label{tab:det}
\vspace{-8pt}
\end{table}

\begin{figure}[t]
    \centering
    \includegraphics[width=0.9\linewidth]{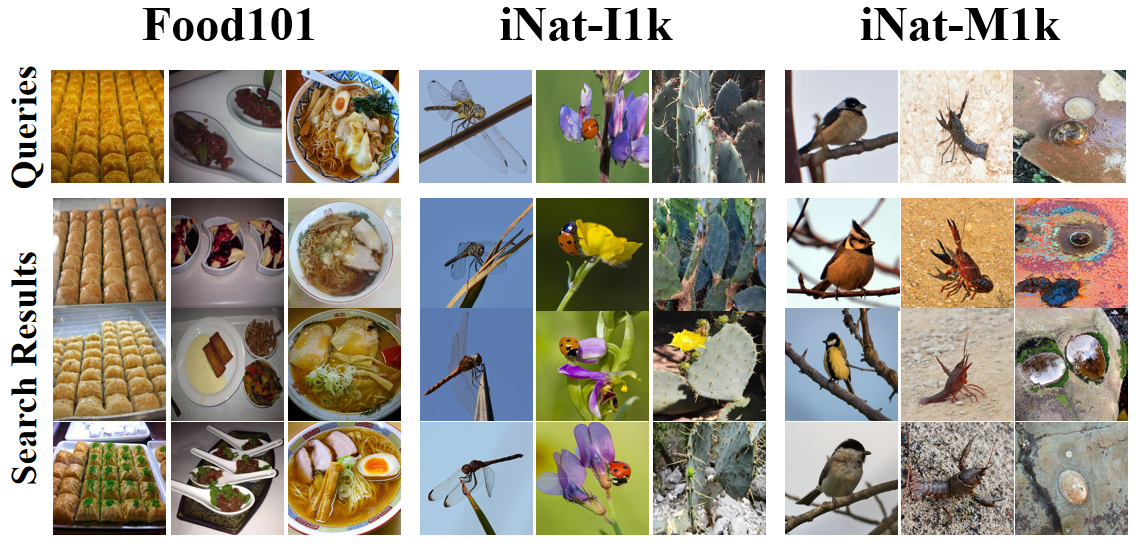}
    \vspace{-8pt}
    \caption{Top-3 retrieval results from 150 million unlabeled images. Query samples from Food101, iNat-I1k and iNat-M1k.}
    \label{fig:retreival_images}
    \vspace{-8pt}
\end{figure}

\vspace{-5pt}
\subsubsection{Visualization of Retrieval samples.}
Fig.~\ref{fig:retreival_images} shows some queries from datasets Food101, iNat-I1k and iNat-M1k and top-3 retrieval results.
We can observe that the retrieval results do not strictly share the same categories with the queries but are visually similar.
In addition, some retrieval results can be dominated by the background or texture.
This indicates that SEPT also will introduce certain task-specific background information.

\section{Conclusion}
\label{sec:conclusion}
In this work, we revisited the scalability and efficiency of transfer learning in the context of scaling up the pre-training datasets. 
We proposed a pre-training framework, SEPT, which takes full advantage of the enormous scale of datasets without prohibitively expensive annotations by selecting the task-specific subset to perform efficiently pre-training via the similarity search.
SEPT only conducts self-supervised pre-training on the retrieval data; thus, it leaves much space for future work to design a more effective algorithm to boost downstream tasks' performance with the retrieval data.
It could also be extended to more different tasks, \eg, segmentation, and more modalities, \eg, vision-language. 
Finally, we hope our work could inspire more research about pre-training from the data perspective for the community.

\vspace{-10pt}
\section{Acknowledgments}
This work was supported partially by National Natural Science Foundation of China (No. 42201358).

\bibliography{aaai23}

\end{document}